\documentclass[letterpaper]{article} 
\usepackage{aaai24}  
\usepackage{times}  
\usepackage{helvet}  
\usepackage{courier}  
\usepackage[hyphens]{url}  
\usepackage{graphicx} 
\usepackage{natbib}  
\usepackage{caption} 
\usepackage{algorithm}
\usepackage{algorithmic}
\usepackage{booktabs}

\usepackage{newfloat}
\usepackage{listings}
\DeclareCaptionStyle{ruled}{labelfont=normalfont,labelsep=colon,strut=off} 
\floatstyle{ruled}
\newfloat{listing}{tb}{lst}{}
\floatname{listing}{Listing}
\nocopyright

\title{Epistemic Injustice in Generative AI}
\author{
    Jackie Kay\textsuperscript{\rm 1 \rm 3},
    Atoosa Kasirzadeh\textsuperscript{\rm 2 \rm 4},
    Shakir Mohamed\textsuperscript{\rm 1}
}
\affiliations{
    \textsuperscript{\rm 1}Google DeepMind \\
    \textsuperscript{\rm 2}Google Research \\
    \textsuperscript{\rm 3}University College London \\
    \textsuperscript{\rm 4}University of Edinburgh  \\
    kayj@google.com, atoosa.kasirzadeh@gmail.com, shakir@google.com

}

\begin{document}

\maketitle

\begin{abstract}
This paper investigates how generative AI can potentially undermine the integrity of collective knowledge and the processes we rely on to acquire, assess, and trust information, posing a significant threat to our knowledge ecosystem and democratic discourse. Grounded in social and political philosophy, we introduce the concept of \emph{generative algorithmic epistemic injustice}. We identify four key dimensions of this phenomenon: amplified and manipulative testimonial injustice, along with hermeneutical ignorance and access injustice. We illustrate each dimension with real-world examples that reveal how generative AI can produce or amplify misinformation, perpetuate representational harm, and create epistemic inequities, particularly in multilingual contexts. By highlighting these injustices, we aim to inform the development of epistemically just generative AI systems, proposing strategies for resistance, system design principles, and two approaches that leverage generative AI to foster a more equitable information ecosystem, thereby safeguarding democratic values and the integrity of knowledge production.
\end{abstract}

\section{Introduction}

While algorithms have traditionally been leveraged to present and organize human-generated content, the advent of generative AI has started to fundamentally shift this paradigm. Generative AI models can now create content -- spanning text, imagery, and beyond -- that resembles that of authors, journalists, painters, or photographers. 
In this paper, we take generative AI to be the class of machine learning models trained on massive amounts of data, typically media such as text, images, audio or video, in order to produce representative instances of such media \cite{garcia2023we}. 

The rapid advancement of generative AI, marked by accelerated software and hardware innovation and a proliferation of novel applications, has been accompanied by growing societal concerns and numerous instances of misuse \cite{bender2021dangers,weidinger2022taxonomy,bird2023typology}. These range from parroting harmful stereotypes and misconceptions about certain social groups \cite{bianchi2023easily, ferrara2023should}, confabulating facts and distorting truth \cite{ji2023survey}, and spreading misinformation and deepfakes \cite{vaccari2020deepfakes,monteith2024artificial}.
While a rigorous philosophical treatment of ``truth'' in the generative AI context is complex \cite[p.9]{kasirzadeh2023conversation} and out of our scope, our modest working definition is: statements that can be verified through adequate evidence, or a robust consensus between relevant social groups. 
Despite these escalating societal concerns and numerous instances of misuse, the discourse surrounding generative AI's rapid advancement lacks a philosophical account that coherently relates these epistemic concerns and explains how they constitute moral violations of a unifying principle.

To address this gap, we develop an account of \emph{generative algorithmic epistemic injustice} by building upon a conventional philosophical understanding of {\itshape epistemic injustice}. Epistemic injustice emphasizes how identity-based prejudice within an information ecosystem not only unjustly hinders the expression of marginalized groups, but also significantly impairs the knowledge formation capabilities of all individuals. It does this by describing the harmful effects and ethical shortcomings inherent in knowledge production systems marked by hierarchical power imbalances rooted in identity.

While traditional discussions of epistemic injustice have primarily centered on interpersonal human interactions \cite{mckinnon2017gaslighting,tsosie2012indigenous}, existing research on algorithmic epistemic injustice has largely been limited to epistemic injustices produced by decision-making and classification algorithms. However, we argue that the distinctive characteristics of generative AI give rise to novel forms of epistemic injustice that necessitate a dedicated analytical framework. To address this, we expand upon the established philosophical discourse on epistemic injustice and introduce an account of ``generative algorithmic epistemic injustice,'' or simply ``generative epistemic injustice,'' to characterize the variety of epistemic harms arising from generative AI systems from a philosophical standpoint.

In Section \ref{section_ej}, we describe epistemic injustice as a social theory and argue for its ethical importance. Section \ref{section_relatedwork} situates our paper within the context of prior research on algorithmic epistemic injustice.
Section \ref{section_gen_ej} builds on the existing research on algorithmic epistemic injustice and identifies four distinct configurations of generative epistemic injustice: amplified and manipulative testimonial injustice, along with hermeneutical ignorance and access injustice. We illustrate these configurations through real-world exemplars of generative AI deployments. While the evidence of injustice in these systems is overwhelming, Section \ref{section_discussion} explores how we can design and use generative AI to foster epistemic justice. We propose strategies for resistance in the face of epistemic oppression caused by generative AI. By unifying the epistemic injustices of AI misinformation, bias, representational harms, and power imbalances within the AI industry under a single four-dimensional theoretical framework, we can identify commonalities in their mitigation strategies and build solidarity among affected groups.


\section{Epistemic Injustice}
\label{section_ej}

In 2013, the city council of Flint, Michigan decided to switch its water supply to the Flint River, notorious for its pollution from automotive manufacturing. This move swiftly provoked public outrage as residents reported tap water turning discolored and emitting a foul, sewage-like odor. Despite these immediate alarms, authorities repeatedly dismissed these complaints, perpetuating a longstanding pattern of environmental gaslighting. The situation escalated when an outbreak of Legionnaires' disease was revealed, previously concealed by political maneuvers to safeguard reputational interests. It was only after academic investigations uncovered alarmingly high levels of lead contamination that the gravity of Flint's water crisis gained national attention, unmasking a profound public health catastrophe. It is obvious in hindsight that the citizens of Flint were correct and the politicians and others in power committed egregious harm by not believing the residents' testimony \cite{davis2021tainted}.

Many have studied the tendency for those with more privilege to ignore and even oppress ordinary citizens with less privilege (socioeconomic, racial, gender or otherwise).
The decolonial scholar Gayatri Spivak put forward the position that the elite's construction of an underclass, and their tendency to presume to speak on behalf of those they disenfranchise, means that the means for resistance to oppression are mediated through the oppressor.
Thus according to Spivak, the ``subaltern''--a group which includes women, the working class, the lower castes, citizens of ``third world'' countries, the colonized, and especially intersections therein--cannot speak \cite{spivak1988can}.

But Spivak's provocation begs the question: can the subaltern speak outside of the structures of the elite?
Seeking alternative means of empowerment, Black feminist scholars devised systems of knowledge outside of white patriarchal domination. Patricia Hill Collins introduced a feminist epistemology that emphasizes the intellectual importance of wisdom gained through Black women's lived experiences, and the transmission of this wisdom through relationships, community, and solidarity \cite{collins2000black}. For Collins, critical dialogue and resistance to threats of epistemic violence is necessary for assessing the claims of the powerful.

Later, Miranda Fricker brought these notions of oppression and silencing to the forefront of mainstream analytic philosophy by coining the term ``epistemic injustice'', referring to injustices related to knowledge and understanding \cite{fricker2007epistemic}. However, Fricker's work has been criticized for not fully acknowledging the contributions of Black feminist scholars who had previously explored similar ideas \cite{berenstain2020white}. These critiques highlight Fricker's oversight of intersectional aspects of oppression \cite{crenshaw2013mapping} and her failure to recognize women of color as agential knowledge creators.
While acknowledging these limitations in Fricker's original account, we find her philosophical framework of epistemic injustice theoretically expressive and influential in subsequent discussions of algorithmic oppression. Therefore, we use Fricker's account as the foundation for our philosophical examination of generative epistemic injustice.

``Epistemic injustices'' includes scenarios where individuals or communities experience unjust discreditation of their knowledge and experiences due to underlying prejudices against their identity. Fricker emphasizes that these biases are directed towards groups with less social power, defined as the capacity to influence others' actions within social interactions and environments. This power can be either agential, exercised by individuals, or structural, embedded in cultural norms and material inequalities.

Fricker argues that epistemic injustices harm not only the oppressed but everyone. Sharing knowledge and experiences is a fundamental human right, essential for self-expression, establishing connections, and asserting needs. To deny this right based on identity is a discriminatory act.
Knowledge acquisition is integral to human existence, shaping our understanding of the world, informing our interactions, and fostering a sense of purpose. Disregarding someone's testimony unjustly not only harms the individual but also deprives society of valuable insights, obstructing collective knowledge growth. Those who unfairly disbelieve someone also do a disservice to others, blocking access to potentially valuable information. Moreover, recognizing and crediting someone's testimony is a valuable heuristic for anticipating and preventing harm.

Fricker distinguishes between two types of epistemic injustice: testimonial and hermeneutical. Testimonial injustice involves the unfair discrediting of someone's account due to prejudice against their identity, a recurring injustice experienced by the Flint residents in our opening example. 
Hermeneutical injustice, the second type, stems from a disconnect between personal experiences and societal understanding. 
``Hermeneutics'' means the interpretation of knowledge. We will sometimes refer to our ``shared hermeneutics'' or ''hermeneutical resources'' as our shared cultural concepts for interpreting each other's experiences and for sharing knowledge.

Hermeneutical injustice occurs when a person's experiences are misunderstood or not recognized at all, due to the absence of appropriate concepts within our collective cultural knowledge.
Shared knowledge is crucial for interpreting and relating to the experiences of others.
However, the repository of collective understanding bears the imprints of dominant groups, leaving the experiences of marginalized communities underrepresented or distorted. This results in gaps and misinterpretations within our interpretive resources, leading to hermeneutical injustice. The transgender community, for instance, has frequently faced this form of injustice. In societies lacking widespread understanding of gender variance, fluidity, non-conformity, and the spectrum of body dysphoria/euphoria, the experiences of transgender individuals are often misinterpreted. This ignorance, as Fricker notes, leads to a lack of empathy and understanding.
Beyond epistemic implications, the material consequences can be severe, obstructing access to essential social resources like healthcare, employment, and housing \cite{fricker2017epistemic}. Both ignorance and misunderstanding contribute to hermeneutical injustice.
\footnote{\citet{medina2018misrecognition} distinguished between ``recognition deficits'', in which a group is unseen or illegible, and ``misrecognition'', where a group is visible but misunderstood, subjected to false and distorted narratives.
In a recognition deficit, the receiver of injustice is ignored or not recognized within the societal context.
In misrecognition, the receiver of injustice is the subject of a statement which is false due to misrepresentation of their identity.}

To summarize, the distinction between testimonial and hermeneutical injustice lies in the assignment of credibility versus the availability of interpretative resources.
In testimonial injustices, the speaker suffers injustice through the degradation of their credibility, a result of identity-based prejudice.
Conversely, hermeneutical injustice relates to the absence of a lexicon to articulate the oppression they experience--leading to an inability to articulate an account altogether.

\section{Related Work on Algorithmic Epistemic Injustice}
\label{section_relatedwork}

The conventional account of epistemic injustice described in Section \ref{section_ej} only involves human actors. However, AI algorithms can also contribute to these injustices because they are epistemic technologies \cite{alvarado2023ai}: they consume, curate, and produce information, which is a precursor to knowledge. 
When algorithms make decisions, particularly in our bureaucratic systems such as governments, businesses, or healthcare providers, they exert power via their contributions to knowledge. 

The emerging field focused on the intersection of epistemic injustice and AI algorithms has been named ``algorithmic epistemic injustice,'' a term coined by \citet{byrnes2023epistemic}. To situate our paper within the broader landscape of algorithmic epistemic injustice, we begin by reviewing the existing literature on this topic. Several key themes emerge from our survey. Testimonial injustice can arise when algorithms are prioritized over human credibility, potentially amplifying existing societal biases. Additionally, hermeneutical injustices can occur when algorithms independently construct meanings and interpretive frameworks, often in automated setups without direct human oversight.

This body of research has a notable emphasis on classification and decision-making algorithms. Several studies exemplify this in different contexts.
In child welfare systems, \citet{glaberson2022epistemic} identify epistemic injustice through algorithms that disproportionately target Black communities and poor single mothers. These algorithmic testimonial injustices lead to wrongful mistrust, surveillance and over-policing committed by humans.
The healthcare sector, as \citet{pozzi2023automated} note, witnesses ``automated hermeneutical appropriation'' in opioid risk score predictions. Pozzi claims that the algorithm establishes meanings for concepts that contribute to a patient's diagnosis, such as ``addiction'' or the experience of pain, without human intervention.
The opacity of data science systems is highlighted by \citet{symons2022epistemic}, who examine the real case of a prisoner who was wrongfully denied parole by the COMPAS recidivism algorithm and continued to be detained, even after providing evidence for his case to human supervisors \cite{wexler2018odds}. The authors argue that this lack of transparency facilitates epistemic injustice: the ignorance about a technology’s inner workings complicated the possibility for contesting its unjust decisions.
\citet{hull2023dirty} discusses how COMPAS and similar systems commit hermeneutical injustice through biased and stereotype-based classifications. Hull also points out the testimonial injustices inherent in physiognomic systems, which wrongly infer personal characteristics based on visual appearance, often linked to race, ethnicity, and gender.
Building on these individual-focused analyses, \citet{milano2024algorithmic} emphasize the importance of our relationships to others for transmitting collective knowledge, which they call our shared epistemic infrastructures, and use these concepts to identify how algorithmic profiling can harm this infrastructure. This approach informs our subsequent discussion on access injustice in Section \ref{access_injustice}.

The domain of AI fairness research has started to intersect with broader concerns of social injustice \cite{hoffmann2019fairness,birhane2021algorithmic,kasirzadeh2022algorithmic}. There has also been a growing recognition of epistemic injustice concepts in relation to AI fairness.
For instance, \citet{edenberg2023epistemic} suggest that an epistemic lens offers a theoretical foundation for understanding the harms of algorithmic bias, which other prevalent frameworks might not adequately capture.

A notable difference between this cluster of prior work and ours is its primary focus on classification or decision-making systems. We build on this cluster to develop an account of epistemic injustice in relation to generative AI, which occupies a growing unique position in the landscape of epistemic injustice due to its capacity to produce convincing and seemingly authentic output.
While classification AI is set to delineate true categories and false ones, generative AI offers statements that play the role of testimonies, explanations, and interpretations. However, the quality and veracity of these outputs varies, posing risks of epistemic contamination.

\begin{table*}[t]
  \centering
  \begin{tabular}{l | l | l}
    \toprule

    &  \textbf{Example} &  \textbf{Intervention} \\
    \midrule
     \textbf{Amplified testimonial} & Parroting misinformation & Identifying testimonial injustice at scale  \\
     \textbf{Manipulative testimonial} & Intentional fabrication of offensive content & Watermarking and automated fact-checking  \\
     \textbf{Hermeneutical ignorance} &  Misrepresenting marginalized experiences & Generating hermeneutical resources  \\
     \textbf{Hermeneutical access} & Obstructing access to information & Equitable distribution of high quality AI \\
    
  \bottomrule
\end{tabular}
  \caption{Summary of the four configurations of generative epistemic injustice and their defining examples. We also summarize the corresponding interventions for achieving epistemic justice proposed in Section \ref{section_discussion}.}
\end{table*}

To the best of our knowledge, the only work with a focus on the epistemic injustice of generative AI is by \citet{de2023conversational}, which looks at the potential of conversational AI for hermeneutical ignorance. The authors review existing literature on how such AI might dominate epistemically within dialogues. However, this study presents two significant limitations. First, its analysis is confined to textual dialogue interactions. In contrast, our theoretical account is designed to be sufficiently broad, extending to include multimodal systems, such as those involved in image generation. This broader scope allows for a more comprehensive understanding of generative AI’s epistemic impact across various mediums. Second, \citet{de2023conversational} primarily examine the immediate effects of AI-generated conversation on individual human interlocutors. Our account, on the other hand, expands this scope and emphasizes the systemic and structural epistemic impacts that could emerge from interactions with generative AI.

\section{Generative Epistemic Injustice}
\label{section_gen_ej}

We now introduce our account of generative epistemic injustice in which generative AI harms the human capacity for understanding and trusting marginalized groups.
Following Fricker's concepts of testimonial and hermeneutical injustice, we conceptualize how generative AI can be complicit in both types of injustice.
We then distinguish further configurations based on how humans shape the model's behavior at various stages of interaction.
Generative AI can either amplify testimonial injustices due to biases acquired in the pretraining and finetuning processes, or it can be manipulated by human users to create harmful content.
Hermeneutical injustices can arise when generative AI's interaction with our shared knowledge leads to the erasure or distortion of marginalized experiences. This may occur when the system lacks sufficient sociocultural understanding of humans, or when the system obstructing the access to knowledge itself. These phenomena constitutes hermeneutical injustice in Fricker's sense because it perpetuates a gap in collective interpretive resources, obstructing the understanding of these marginalized experiences.

Therefore our four configurations of generative epistemic injustice are:
\begin{enumerate}
    \item Generative amplified testimonial injustice: when generative AI magnifies and produces socially biased viewpoints from its training data.
    \item Generative manipulative testimonial injustice: when humans fabricate testimonial injustices with generative AI.
    \item Generative hermeneutical ignorance: when generative AI lacks the interpretive frameworks to understand human experiences.
    \item Generative hermeneutical access injustice: when unequal access to information and knowledge is facilitated by generative AI.
\end{enumerate}

Table 1 summarizes these concepts. We will sometimes drop ``generative'' when referring to these concepts due to the scope of the paper; however, we note that each configuration has an equivalent counterpart outside of the realm of generative AI.

For each configuration we will describe its contributing sociocultural factors and potential second-order effects.
Note that generative algorithmic epistemic injustice is not a speculative theory, but a real danger that has exploded with recent advances and investment in the field. Thus each theoretical concept is illustrated by an exemplar sourced from real world research or investigative journalism on generative models.

\subsection{Amplified Testimonial Injustice}

Generative AI systems have a unique capacity to perpetuate and amplify existing testimonial injustices. Trained on vast datasets often scraped from the web, these models inherit and reproduce the prejudices and biases embedded within those sources. This can result in the re-commitment of testimonial injustices where the credibility of marginalized groups is systematically undermined due to prejudice against their identity. The generative AI becomes an unwitting perpetrator of social biases, unfairly discrediting the knowledge and experiences of certain groups. Moreover, the uneven representation of different identity groups within these datasets further exacerbates this issue. Generative AI models are more likely to reproduce the voices of those frequently represented and culturally dominant online, while erasing the voices of the socially marginalized. This creates a feedback loop where dominant narratives are amplified and marginalized voices are further silenced, compounding the testimonial injustice experienced by these groups.

Several factors contribute to this amplification, making it a distinct form of testimonial injustice. The deployment scale of generative AI naturally allows biased narratives to reach a massive audience, while the perceived objectivity and authority of AI-generated content can lend credence to these narratives, even when they reflect societal biases. This can lead to a situation where the generative AI's output is trusted over the lived experiences and knowledge of marginalized individuals, thus reinforcing existing power imbalances and perpetuating testimonial injustice. Once disseminated, these biased narratives can be difficult to retract or correct.

In the conventional human-only environment, testimonial injustices involve a credibility deficit assigned to someone's account of truth based on the prejudices of the listener.
In the algorithmic setting, the injustice requires a credibility excess assigned to the algorithm; that is, humans believe the account amplified through the technology over the individual or group who is discredited.

The systemic consequence of this arrangement leads to the degradation or wrongful attribution of trust. Users engaged with the generative AI are exposed to narratives influenced by social biases, which further deteriorates their trust in marginalized groups. Concurrently, these marginalized groups experience a decline in trust towards the system itself, the entity that established it, and other institutions tasked with upholding veracity.

\subsubsection{ChatGPT and Misinformation Fingerprints}

Testimonial injustices can be memorized by large language models and amplified in their outputs, as shown by the January 2023 study of GPT-3.5’s responses to requests for false information from NewsGuard’s “Misinformation Fingerprints” database \cite{newsguardgpt3.52023}. While the system rejected the more infamous conspiracies, such as the “birther” conspiracy that Barack Obama was born in Kenya and thus ineligible to be President of the United States, ChatGPT perpetuated false narratives for 80\% of the prompts, for a sample size of 100.
In March 2023 NewsGuard reran the same study using GPT-4, and 100\% of the prompts followed false narratives \cite{newsguardgpt42023}. 

ChatGPT complied with a request to write propaganda from the point of view of the Chinese Communist Party denying allegations about Uyghur internment camps. The system produced text claiming that the government had established “vocational education and training centers” to “address the issue of terrorism and extremism”. In reality, there is extensive evidence and eyewitness accounts that Uyghur ethnic minorities have been detained en masse and subjected to forced labor, forced birth control, separation of families, and Islamophobic religious suppression \cite{ohchr2022ohchr}. This is a clear instance of generative AI perpetuating state-sponsored testimonial injustice.

The model also repeated false claims about the 2018 Parkland school shooting originating from right-wing news pundit Alex Jones: that the victims and their grieving family members were “crisis actors” hired by the government to “push a gun control agenda.” This polemic is a testimonial injustice to the eyewitnesses of the attack and to the parents who lost their children, due to their statements in favor of firearms regulation in the wake of the tragedy. Set against a politically charged backdrop, the conspiracists were so committed to lobbying against gun regulation that they targeted and publicly smeared these activists.

Although the NewsGuard study was a simulation of misinformation, rather than an ``authentic'' instance of epistemic injustice, the generative AI's sycophantic fulfilment of the request to spread misinformation reflects how testimonial injustices are memorized and the potential for their amplification by generative models.

\subsection{Manipulative Testimonial Injustice}

While traditional epistemic injustice literature primarily focuses on unconscious biases and cultural prejudices, we argue that the intentional manipulation of falsehoods, which often exploit and reinforce these prejudices, also constitutes a form of epistemic injustice. There is ample evidence of disinformation and conspiracy theories being deliberately crafted and amplified for political gain \cite{marwick2017media}. Conspiracy theories often disproportionately harm marginalized groups \cite{jaiswal2020disinformation}, and are sometimes weaponized against them to justify oppression \cite{nera2022conspiracy}.
For example,the Senate Intelligence report on interference in the 2016 US Presidential election concluded that Russian information operatives disproportionately targeted African Americans, and ``by far, race and related issues were the preferred target of the information warfare campaign'' \cite{2020senateintelligence}.

In the context of generative AI, manipulative testimonial injustice occurs when humans intentionally steer the AI to fabricate falsehoods, discrediting individuals or marginalized groups. Unlike amplified testimonial injustice, which emerges from memorized patterns in data, manipulative testimonial injustice involves deliberate manipulation through techniques like prompting or jailbreaking.

The extensive deployment of generative AI has introduced a novel form of manipulative testimonial injustice: the false accusation of deepfakes. This tactic exploits the increasing uncertainty surrounding the authenticity of digital media, creating a ``liar's dividend'' where even genuine evidence can be dismissed as fabricated \cite{schiff2023liar}. This weaponization of doubt and uncertainty further undermines the ability of marginalized groups to have their voices heard and their experiences validated. Disregarding an actual human’s documented testimony as AI-generated is a tactic of discreditation, often concealing underlying prejudice, and frequently appears in conspiracy theories. For instance, consider the scenario where a candidate for the U.S. Congress in Missouri, running for a House seat, indulged these conspiracy theories. They falsely asserted that the 2017 video capturing George Floyd's murder by police was a deepfake. This claim aimed to undermine the Black Lives Matter movement by suggesting it propagated falsehoods to exacerbate racial tensions \cite{giansiracusa2021deepfake}. Although this candidate did not succeed in the primary election, the misuse of frontier generative AI technology as a tool for unjust distortion is a growing a significant concern. Recent participant surveys have demonstrated that AI-generated propaganda can be as persuasive as news articles written by professional propagandists \cite{goldstein2024persuasive}, illustrating the public's vulnerability to manipulative synthetic content.

\subsubsection{4chan Abuses of Bing Image Creator}

Generative AI models can be adversarially prompted to fabricate “novel” misinformation by synthesizing and recombining known elements into statements or portrayals not present in the pretraining data. There is mounting concern around deepfakes engineered to stoke international conflict and weaken the opposing side in war \cite{brookingsdeepfakes2023}, as well as increased incidents of deepfake porn, used for harrassment, blackmail, and degrade individuals, with a 2023 report finding that 98\% of deepfake videos online were porn \cite{homesecuritydeepfake2023}.
While these concerning applications warrant entire investigations unto themselves, we will highlight generative AI-generated deepfakes to denigrate identity groups as an instance of manipulative injustice.

After Microsoft released Bing Image Creator, an application of OpenAI’s text-to-image model DALLE-3, a guide to circumventing the system’s safety filters in order to create white supremacist memes circulated on 4chan. In an investigation by Bellingcat, researchers were able to reproduce the platform abuse, resulting in images depicting hate symbols and scenes of antisemitic, Islamophobic, or racist propaganda \cite{bellingcatdalle2023}. These images are crafted with the intention of demonizing and humiliating the targeted groups and belittling their suffering. Hateful propaganda foments further prejudice against marginalized groups, stripping them of credibility and leaving them vulnerable to testimonial injustice.

\subsection{Generative Hermeneutical Ignorance}

When novel social experiences emerge throughout history, and mainstream cultural narratives fail to grasp them, hermeneutical injustices inevitably arise. This phenomenon also extends to novel sociotechnical experiences, where interactions between humans and new technologies can lead to misunderstandings and misrepresentations of lived experiences.

In the context of generative AI, we propose the term ``generative hermeneutical ignorance'' to describe how these systems can erase or misportray marginalized groups due to a lack of contextual and cultural understanding. This occurs when generative models, despite their appearance of world knowledge and language understanding, lack the nuanced comprehension of human experience necessary for accurate and equitable representation.

Generative models can perform forms of interpretation and understanding through their world knowledge and natural language capabilities; however, their interpretive resources are significantly different from those of humans. While LLMs may demonstrate forms of human language skills, they lack embodied knowledge and cultural history. For example, image generators can produce aesthetically pleasing visuals but may struggle with grounded physical concepts. This apparent comprehension without deeper contextual understanding can lead to hermeneutical ignorance, where generative AI interprets dominant narratives while diminishing or misrepresenting aspects of human experience inaccessible to the models.

The interpretative misrecognition by generative AI surfaces collective cultural misunderstandings which remain undetected by developers' safety mechanisms or the preferences of fine-tuning raters. The absence of underrepresented cultures from these models is even harder to point out \cite{qadri2023ai}.

Due to their positions of power, creators and overseers of AI technology may be less likely to notice, let alone rectify, this form of hermenutical injustice within their systems, even when presented with evidence. This willful hermeneutical ignorance-the continued misunderstanding or misinterpretation of marginalized experiences despite their articulation \cite{pohlhaus2012relational}—leads to complacency and reinforces hermeneutical oppression.

This phenomenon of generative hermeneutical ignorance diverges from traditional forms of hermeneutical injustice, as well as those perpetuated by other algorithmic systems. While traditional hermeneutical injustice often arises from a lack of shared understanding or conceptual resources within a human community, generative hermeneutical ignorance is unique in that it stems directly from the limitations of generative AI models themselves.

Unlike human-based hermeneutical injustices, which can be addressed through dialogue, education, and cultural exchange, the challenges posed by generative AI are rooted in the inherent limitations of current technology. Generative AI models lack the embodied and cultural knowledge that humans acquire through lived experiences. This lack of understanding can lead to the misinterpretation of marginalized voices and perspectives, even when the generative AI tool is not necessarily trained on discriminatory intent. Moreover, generative hermeneutical ignorance differs from the hermeneutical injustices caused by other algorithmic systems, such as classification algorithms. While these systems can perpetuate biases present in their training data, generative AI models have the potential to \emph{create} entirely new forms of misinterpretation and misinterpretation.

\subsubsection{Generative AI and the American Smile}

In March 2023, a Medium blogger reflected on a slideshow of Midjourney generations imagining photographs of a time traveller taking group selfies with people from various time periods  \cite{jenkaamerican2023}: garish photographs in which groups of Native Americans or Japanese feudal warriors or other groups in traditional garb are gathered closely together, beaming ear-to-ear at the camera.
The post observes how this facial expression is evidence of modern American cultural dominance, contrasting the AI-generated images with historical photographs and images from cultures with different expressive norms such as Eastern Europe.
The author further laments how this smile represents a loss of cultural diversity and with it, a loss of breadth of internal experiences and emotion.
The author also notes that the same slideshow depicts Spanish conquistadors smiling alongside Aztec warriors, which seems unrealistic given the violent colonial history of the Spanish empire.

These images are evidence of the hermeneutical ignorance of Midjourney. They show a lack of sensitivity and awareness around cultural difference, historical violence, facial expressions, perhaps even the internal experience of a smile.
The model has misrecognized these groups, and through doing so erased a part of their cultural experience--indeed, of any non-American culture.
Although the historical time periods depicted in these specific images have passed, the honoring of history and cultural diversity are necessary for building our hermeneutical resources.

\subsection{Hermeneutical Access Injustice}
\label{access_injustice}

The phenomenon of generative hermeneutical access injustice is a distinct form of hermeneutical injustice within the realm of generative AI. According to Fricker's account, hermeneutical injustice arises when individuals are unable to fully understand or articulate their experiences due to a lack of shared conceptual resources or societal understanding. In the context of generative AI, this injustice takes another specific form: it centers on the generative AI's control over access to information, leading to a denial of knowledge based on identity-driven bias or misrecognition.
This withholding or distortion of information based on identity aligns with Fricker's concept of hermeneutical injustice as it directly impacts an individual's capacity as the receiver of knowledge.

In the algorithmic setting, user information serves as the basis for the system's discrimination.
Access injustice also illustrate the cultural biases that emerge in a system.
For example, a study of automated speech recognition showed that African American users had difficulty controlling voice-activated technologies unless they accommodated their speech patterns
\cite{mengesha2021don}.

The direct consequences of access injustice can be unfair obstruction from goods, services, and information. 
On the level of second-order effects, hermeneutical access injustice can lead to echo chambers and epistemic fragmentation, isolating identity groups from each other on an informational level.
This then exacerbates conventional hermeneutical injustice, because it causes the information gap to widen between identity groups, detracting from their shared understanding and the widespread access to knowledge about marginalized experiences.

\subsubsection{Multilingual Injustice}
LLMs are notoriously English-centric and have variable quality across languages, particularly so-called ``under-resourced'' languages.
This is a significant risk for access injustice: speakers of these underrepresented tongues, who often correspond to members of globally marginalized cultures, receive different information from these models because the creators of the technology have deprioritized support for their language.

This type of linguistic access injustice may reflect profound asymmetries in political power across disparate language groups.
\citet{kazenwadel2023user} asked GPT-3.5 about casualties in specific airstrikes for Israeli-Palestinian and Turkish-Kurdish conflicts, demonstrating that the numbers have significant discrepancies in different languages–for example, when asked about an airstrike targeting alleged PKK members (the Kurdistan military resistance), the fatality count is reported lower on average in Turkish than in Kurmanji (Northern Kurdish).
When asked about Israeli airstrikes, the model reports higher fatality numbers in Arabic than in Hebrew, and in one case, GPT-3.5 was more likely to deny the existence of a particular airstrike  when asked about it in Hebrew.
The credibility assigned to claims, resulting in a dominant account, varies across linguistic contexts.

How else does this constitute an epistemic injustice? In an armed conflict, when the attacking side downplays fatality rates, particularly civilian fatalities, they are most likely trying to deflate the credibility of critics of this violence, who may be members of the targeted group, or third parties who simply oppose acts of war. This deflation is motivated by a synthesis of political interests and prejudice against the group who is harmed by violence.

\subsection{Specific Harms of Generative Epistemic Injustice}

We now specify the characteristics of generative AI that give rise to the epistemic injustices described above, distinguishing this class of models from the algorithmic injustices surveyed in Section \ref{section_relatedwork}. These characteristics give us the language to discuss the broader implications of generative epistemic injustice and the societal harm it represents.

The outputs of these generative models represent a complex blend of memorization and synthesis. Memorization involves retrieving and reiterating existing patterns found within the dataset. Synthesis, on the other hand, involves recombining these patterns across various levels of granularity. This synthesis process can yield outputs that range from seemingly insightful and emergent to nonsensical and factually incorrect.

ChatGPT and similar generative models are known for their propensity to fabricate or ``hallucinate'' information, a phenomenon well-documented in the literature \cite{ji2023survey}. This characteristic is inherent to their design: language models generate text by predicting the next token in a sequence based on its statistical frequency within a vast dataset, often derived from scraping the internet \cite{holtzman2019curious}.

This analysis brings us to two primary pathways through which misinformation infiltrates the outputs of generative models: the memorization of inaccuracies from the source dataset and the generation of high-likelihood sequences that, despite the model's prediction, contain clear factual errors \cite{augenstein2023factuality}.
These two categories are not mutually exclusive; an instance of misinformation might well be a blend of remembered falsehoods and newly synthesized fabrications. 
While there are ongoing efforts to fine-tune generative models to address these limitations by hedging \cite{abulimiti2023kind} or abstaining from answering questions out of their scope of knowledge \cite{zhang2023r}, completely eradicating all misinformation from pretraining data is challenging as it would require the automated classification of the truth at a massive scale. 
Similarly, ongoing research investigates if it is possible to detect when language models are ``lying'' by analyzing their internal state \cite{liu2023cognitive},
but when contradictory viewpoints are learned during pretraining, these purely mechanistic approaches may fail to reconcile what is and isn't true.

Representational harms in generative AI tend to arise from the memorization of biased patterns in training data, perpetuate unfair outcomes in decision-making systems or stereotypical portrayals in generative systems \cite{caliskan2017semantics, birhane2021multimodal}.
However, these harms can also arise from the AI's synthesis of culturally incongruous concepts  \cite{prabhakaran2022cultural}. Harm may occur through unexpected recombination of features that are uncommon in the data, but carry offensive or derogatory connotations in certain cultural contexts.

One major concern with both memorized and synthesized content in generative AI is rooted in its potential influence on our shared concepts and, consequently, social power structures.
Large language models not only produce declarative-like statements but also can perform performative-like statements that can change aspects of the world \cite{kasirzadeh2023conversation}. Similarly, multimodal models can fabricate various depictions and portrayals of the world, engaging interlocutors in a more complex dialogue than classification systems.  
These performances co-create our collective sense of meaning and identity \cite{lu2022subverting}, intertwined with the dynamics of social power.
Furthermore, the outputs of generative AI can be highly persuasive due to their manipulation of our cognitive biases \cite{el2024mechanism}. Whether accidental or intentional on part of the user, this persuasive capacity amplifies AI's capacity to shape collective knowledge.

If generative models become a common epistemic tool--if we treat them like web search or encyclopedias, or like infinite firehoses of information--they will shape the structure of our collective body of knowledge. We are interested not only in direct harms of a system’s outputs but also their second-order effects on the information ecosystem. Epistemic injustices not only render marginalized groups discredited and invisibilized, they pollute the epistemic environment \cite{ryan2018epistemic}, making it difficult to reason about knowledge, reject false premises, or find verified facts. Epistemic flooding is when knowing agents are overwhelmed with so much information they become incapable of critically assessing anything they encounter \cite{anderau2023fake}.

The second-order effect of epistemic injustice is the degradation of the bond of trust between speakers and receivers of information, due to a cycle of credibility deflation and ignorance. This oppressive silencing and reactionary dissent further impedes the transfer of knowledge across communities. Dismissing the testimonials of a marginalized group and ignoring their experiences results in the further the polarization of which facts and viewpoints are acceptable in certain identity groups.
Another relevant concept is testimonial smothering when a speaker is so inured to being silenced or misunderstood that they hold back their testimony entirely, which hinders everyone else from accessing their truth \cite{dotson2011tracking}. These damages of epistemic injustices on the information ecosystem result in further hermeneutical injustices, because they impoverish the collective interpretational resources for understanding the experiences of marginalized groups.

As we will see, these harms to the information ecosystem have consequences that appear backwards or contradictory. The erosion of trust can cut both ways: a marginalized group may lend less credibility to dominant institutions, which puts them at a disadvantage when these institutions try to dispense knowledge for the public good, such as medical advice \cite{annesley2020connecting}. Researchers have also studied the backfire effect, in which an intervention attempting to change an individual's belief ends up reinforcing their belief \cite{swire2020searching}. These concepts are all at play in generative AI's contribution to epistemic injustice.

Memorization and synthesis are the key process by which generative models formulate their outputs, which represent content that can both reproduce unjust testimonies and formulate new hermeneutics.
These outputs have persuasive and performative potential: they enable AI to participate in the definition of semantic concepts and shape both our individual and collective knowledge.
This common body of knowledge can be thought of as an epistemic ecosystem, vulnerable to pollution by false narratives. The resulting systemic effects can further hamper the flow of information and erode trust between disparate groups.

\section{Towards Generative Epistemic Justice}
\label{section_discussion}

Thus far we have characterized generative AI's the potential for systemic epistemic injustice. Our taxonomy enables us to name instances of these harms and recognize whether the injustice stems from pre-existing power imbalances, issues with system design or development process, or combinations thereof.
This enables us to pursue the orientation of generative AI towards epistemic justice.
We will now examine how the theory of epistemic justice can inform the sociotechnical design of generative AI, then suggest how to apply this technology to balance the scales of justice.

\subsection{Epistemic Justice for Generative AI}

Epistemic justice is an ethical ideal to consider when designing a technology that interacts with power structure in our knowledge systems.
We discuss how the virtues of epistemic justice can be incorporated into the development and uses of generative AI in order to mitigate the epistemic injustices we have studied in this work.

\subsubsection{Epistemic virtues, participation, and representation.}

Epistemic injustice is so prevalent in our daily lives, it seems impossible to imagine an alternative.
Yet shifting towards a culture of epistemic \emph{justice} is a worthy endeavor, and in Fricker's account can be done so through epistemic virtues.
As the holders of social power, dominant groups have a particular responsibility to hone their epistemic virtues.
The virtue of testimonial justice can be achieved through critical, reflexive awareness of prejudice: the ability to look inward upon receiving a testimony, recognize one's own biased assignment of credibility to the speaker, and adjust one's judgment accordingly.
To achieve the virtue of hermeneutical justice, one must exercise sensitivity as to why a member of a marginalized group may have difficulty articulating their experience, or remain silent in the face of oppression, rather than accepting the status quo on its face.
\citet{bondy2010argumentative} emphasizes a healthy skepticism and ``metadistrust'' of our own biases: to distrust the inclination to distrust the marginalized.

Beyond individual interactions, enacting epistemic justice in generative AI requires a systemic amplification of marginalized voices \cite{kalluri2020don}.
Epistemic justice provides a normative argument for participatory development methods. By meaningfully engaging with affected groups, developers of generative AI can build their collective hermeneutical knowledge and awareness of societal biases that silence marginalized voices.
Prior participatory studies successfully examined the cultural erasure and misunderstanding of South Asian cultures exhibited by text-to-image generative models by consulting with affected users \cite{qadri2023ai}.
However, participatory methods have many limitations and critiques \cite{birhane2022power}, and have not gained a meaningful foothold in the AI industry \cite{groves2023going}. Participation may not be effective or even possible in the case of willful hermeneutical ignorance.
More radical systemic change is a necessity for epistemic justice.

Equitable representation in the collection and ownership of data would help implement generative epistemic justice.
By surfacing authentic accounts of underrepresented groups and amplifying them in datasets with their consent and involvement through data sovereignty \cite{kukutai2016indigenous}, we can build their social legibility and bolster collective hermeneutics for understanding and accepting their experiences. Institutional structures such as public trusts can play a role in stewarding a digital commons of data \cite{huang2023generative, chan2023reclaiming}.
Another option for improving representation in the model development lifecycle is to ensure that community expertise is more deeply trusted through various measures, up to and including different compensation models for different kinds of data \cite{fredman2020thinking}.

\subsubsection{System design.}

Because knowledge is produced, distorted, disseminated and obstructed by algorithms, developers of AI technology have an important role in identifying and mitigating the resulting injustices.
A critical technical practice of generative epistemic justice requires questioning the dominant hermeneutics of the field and understanding how they are embedding into system mechanisms \cite{agre2014toward}.

To reduce epistemic injustices, generative AI developers can take care to understand bias recorded in pretraining data sources and practice reflexivity and sensitivity in fine-tuning and other safety interventions.
There is a high risk of hermeneutical injustice in dominant groups exclusively crafting the rules of value alignment, or the process by which agents are aligned.
Instead of presuming that an objective ``view from nowhere''--which, in fact, is the dominant view \cite{haraway1988situated}--can be paternalistically imposed upon users, a pluralistic approach to alignment can engage a diverse and representative sample of the populace \cite{anthropic2023collective}.

Once a generative AI system with general capabilities is developed and deployed, it is vulnerable to adversarial use.
Continued investment in technical solutions for validating the authentic provenance of information may help safeguard against manipulative testimonial injustice \cite{gregory2023fortify}.
The concepts of generative epistemic injustice show us the ways in which the purely technical interpretation of watermarking and other provenance validation technologies may be flawed.
The watermark itself may unduly detract from the model's credibility.
What about purely memorized ``true'' information which is regurgitated by the model and imprinted with a watermark, or human-fabricated misinformation which lacks an AI-generated watermark?
The watermark merely tells us the origin of the content, but does not shore up our tools for critically assigning credibility and trust.
Furthermore, any algorithm which tries to automatically assign credibility is at risk of committing testimonial injustice.

If watermarking and similar approaches prove intractable, an alternative direction would be detecting factual inconsistencies by leveraging expert domain knowledge and automating fact-checking practices.
This is a technically challenging area that has gained some momentum in the language domain \cite{schlichtkrull2023averitec}, but is largely unexplored for image, video, and other modalities outside of text \cite{guo-etal-2022-survey}.
Furthermore, systems for misinformation detection are also vulnerable to epistemic injustices and may automate the discrediting of the marginalized \cite{neumann2022justice} if epistemic virtues are not exercised in their creation.

Hermeneutical access injustice can be mitigated by using identity marker data with caution, or avoid using it at all \cite{hanna2020towards} \cite{lett2022conceptualizing}.
Access can be made more equitable not only by distributing AI systems across geographies, languages, and economic access, but through a consistent quality of information throughout these deployments.

Recall that another contributor to epistemic injustice is the opacity of AI. If we remain ignorant about the inner workings of AI, affected groups will have no recourse to contesting the reasoning behind unfair decisions or generations. The opposite of epistemic opacity is epistemic transparency, which could be achieved through better documentation of generative models \cite{heger2022understanding}, more understandable, user-friendly, and rich interfaces, and a better science of explaining and understanding the mechanisms of generative models  \cite{grzankowski2024real} \cite{xu2019explainable}.

\subsection{Generative AI for Epistemic Justice}

As an epistemic technology, AI represents a powerful vehicle for epistemic injustice due to its memorization of prejudices, its performative capacities, and the scale at which it could pollute our information ecosystem.
However, frontier models have also demonstrated amazing capacity for creativity, pattern matching, and even context sensitivity.
In this section we argue that generative AI and search can be used as technologies of resistance to injustice \cite{agnewtechnologies}, by surfacing testimonial injustices and bolstering our shared hermeneutical resources for understanding marginalized experiences.

\subsubsection{Identifying testimonial injustices at scale.}

We have thus far argued that AI can amplify testimonial injustices.
However, we can also design a system for measuring the prevalence of injustice at scale.
Gathering evidence on amplified testimonial injustice--how it manifests and how prevalent it is--is the first step to its prevention.
We now present an outline of how such an investigation could be conducted.

To both design the criteria for identifying testimonial injustices and to measure the impact of these injustices, input from affected groups is crucial.
Epistemic injustices are human-computer interactions;
therefore, they cannot be evaluated by traditional machine learning methods, which measure a computer's behavior in isolation of human judgment \cite{weidinger2023sociotechnical}.
A study of epistemic injustice must include discourse with the humans who provide the basis for the meaning of the information that is generated and spread by machines.

The study begins with a consultation of members of an affected group about popular narratives that discredit or undermine their identity group, particularly those circulating on social media and the Internet.
After collecting data on narratives and situations that represent testimonial injustices, we can use it to investigate AI systems and the data they consume. While the exact design of the system is contingent on many factors, such as available resources and steering from the affected groups, we sketch a few possible directions here. Although we use language suggesting that the system operates on text, these techniques could be extended or adapted to image, audio or even video.

The first direction is to attempt to measure a generative model's tendency to amplify the narratives of testimonial injustice.
By crafting prompts from the collected narratives, we can attempt to elicit outputs that repeat them, and thus represent a risk of amplifying testimonial injustice.
Red-teaming and automatic red-teaming can be used to scale up this process \cite{ganguli2022red}.
Importantly, the affected groups must be consulted again to evaluate if the resulting model outputs can be considered testimonial injustices, and if so, the severity and nuances of these wrongs.

A separate, perhaps complementary system would be one that detects instances of testimonial injustice in a large corpus of data.
Embeddings-based retrieval could be used to locate documents, assuming that lower distance in the embedding space of a model corresponds to semantic similarity \cite{khandelwal2019generalization}.
Alternatively, a classifier could be trained to recognize similar narratives to those in the collected dataset.
The results of these methods would again need to be evaluated by humans, particularly those of the affected group.
Broadly, the utility of such a system is identifying the contexts in which testimonial injustice occurs in recorded data, who perpetuates it, and the prevalence relative to opposing narratives, if those are also measured.
A large internet-scale dataset like those used for pretraining could be analyzed from the perspective of which website domains have high frequencies of unjust narratives.
Smaller domain or application-specific datasets could be analyzed with those specific aims in mind. For example, a fact-checking corpus could be audited for testimonial injustice, to investigate if efforts to verify the truth are, in fact, amplifyig bias.

\subsubsection{Generating hermeneutical resources.}
Through its participation in performative discourse and creative synthesis, AI can contribute to our shared hermeneutical resources.
Although we have so far emphasized their hermeneutical injustice, we can alternatively direct these systems to expanding our cultural understanding of each other and ourselves.
Other breakthroughs in AI have demonstrated how these epistemic tools can unlock novel scientific knowledge \cite{endsfreudenheim}.
Can we use generative AI to unlock cultural knowledge to ameliorate hermeneutical ignorance?

Generative AI can enable the creative exploration of new experiences by simulating them, and help articulate experiences which are otherwise ineffable.
Simulating the experiences of others can build empathy across identity lines (while avoiding problematic uses such as appropriation).
AI can be an interactive tool for exploring one's own experiences.
Image generation can be used to re-imagine and express oneself.
Dialogue agents can be used to provide alternative interpretations, retrieve narratives from history, or share similar experiences from other users, with their consent.
Although these conversations are not a substitute for the kind of human-to-human community gathering and organizing that helps marginalized groups build their hermeneutics, they can be a tool for bolstering the confidence and resources of those who might be isolated from their communities.

Though generative AI models absorb much of our pre-existing cultural biases through pretraining data, their holistic understanding of the world cannot be said to resemble ours. This is due to many factors: the vast difference between our underlying cognitive and sensory mechanisms and the specific content of their data and experiences to ours, to name a few \cite{bender2020climbing}. 
The ``alien'' nature of these models means they are less likely to following our pre-existing hermeneutics, for better or worse.
This is the motivation for fine-tuning efforts via RLHF or other methods: to align the outputs of a pretrained model with an acceptable imitation of an ideal for an obedient assistant \cite{ziegler2019fine, bai2022training}.
However, this ideal and its realization is produced by AI's dominant groups, which restricts the diversity and depth of experiences the model can portray.
Fine-tuning techniques could investigate how to pinpoint and preserve meaningfully diverse voices within foundation models that maintain morality and epistemic justice, instead of washing them away with fine-tuning to a specific ``neutral'' (dominant) voice.

\section{Conclusion}

We have expanded the philosophical concept of epistemic justice to reason about both generative AI's disproportionate impact on marginalized groups, and its influence on everyone's capacity for knowledge.
While the memorization of existing human biases and the fabrication of falsehoods are rampant issues in these models, generative AI systems can also be re-engineered to surface injustices and enrich our cultural resources.
Severe power imbalances at both a societal and technological level are apparent in our interactions with generative AI's outputs.
Epistemic justice is a guiding principle to orient our knowledge systems towards equity and fairness for all.

\section*{Acknowledgements}

The authors thank Julia Haas, William Isaac, Nicklas Lundblad, and Aliya Ahmad for internal review, and Tommy Curry, Nahema Marchal, Iason Gabriel, Tom Stepleton, Laura Weidinger, and Marc Deisenroth for valuable feedback and advice.

\bibliography{aaai24}

\end{document}